%% file: main.tex
\pdfoutput=1
\documentclass[sigconf,nonacm]{acmart}

\setcopyright{none}

\usepackage{amsfonts}
\usepackage{array}
\usepackage{float}
\usepackage{tikz}
\usetikzlibrary{arrows.meta, positioning, calc}
\usepackage[capitalize]{cleveref}
\crefname{figure}{Figure}{Figures}
\Crefname{figure}{Figure}{Figures}
\crefname{table}{Table}{Tables}
\Crefname{table}{Table}{Tables}
\newcommand{\yes}{\ensuremath{\checkmark}}
\begin{document}

\title{Safety-Flag: A Unified Benchmark for the Reliability and Calibration of LLM Content Moderators}

\author{Yibo Hu}
\affiliation{
  \institution{Illinois Institute of Technology}
  \city{Chicago}
  \state{Illinois}
  \country{USA}}
\email{yhu89@illinoistech.edu}

\begin{abstract}
Large language models are increasingly used for content moderation, but most evaluations still report
aggregate accuracy on individual benchmarks. We introduce Safety-Flag, which places seven widely used
safety benchmarks (BeaverTails, XSTest, Ethics, WildGuard, Aegis, ToxiChat, and ToxiGen) into a single
balanced flag / do-not-flag protocol. We release item-level decisions and confidence scores for six
general-purpose LLMs and four dedicated guards, together with three reference models, evaluated on the
same items. Safety-Flag measures three dimensions of moderator reliability: error direction, probability
calibration, and confidence-based error ranking for human review. They often disagree.
Aggregate accuracy does not reveal error direction: one model flags $85\%$ of benign content,
whereas another misses $54\%$ of harmful content. All six general-purpose models are overconfident;
fitting one temperature per model reduces calibration error by $2.8$--$6.0\times$ without changing predicted
labels or confidence ordering. Confidence-based abstention lowers selective risk for every model, although
the gains depend on how well confidence ranks errors. Dedicated guards produce fewer false alarms and are
better calibrated, but several have higher miss rates outside their documented coverage. We release the
benchmark, fixed item lists, evaluation code, per-item model outputs, and leaderboard at:
\url{https://github.com/yibo-hu-lab/safety-flag-benchmark}.
\end{abstract}



\maketitle

\section{Introduction}
\begin{figure}[t]
\centering
\includegraphics[width=\linewidth]{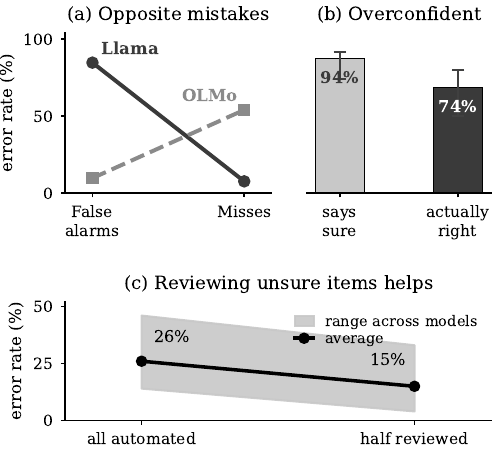}
\caption{\textbf{Aggregate accuracy hides three deployment-relevant differences.} (a)~Llama and OLMo
fail in opposite directions: one over-flags benign content, while the other misses harmful content.
(b)~Averaged across models, stated confidence exceeds actual accuracy. (c)~Reviewing the least-confident
half lowers the error rate on the items that remain automated, although the gain varies across models.}
\label{fig:teaser}
\end{figure}
Large language models (LLMs) are increasingly deployed as content moderators, deciding whether a post,
prompt, or model response should be flagged as unsafe~\cite{inan2023llamaguard,markov2023holistic}. The
decision is consequential in both directions. Over-flagging benign content silences users and overwhelms
human review; missing harmful content defeats the purpose of moderation.

Accuracy and F1 do not answer three questions that matter in deployment. First, does a model mainly
flag benign content or miss harmful content? Second, does its reported confidence match how often it is
correct? Third, when low-confidence items are sent to a human, does the remaining automated error
actually decrease? We call these three dimensions \emph{error direction}, \emph{probability calibration}, and
\emph{error ranking}. \Cref{fig:teaser} shows all three on real moderators: two models fail in opposite
directions, every model is overconfident, and human review helps some models far more than others.
Aggregate accuracy reveals none of this.

Existing evaluations cannot answer these questions together. Safety benchmarks use different labels,
harm taxonomies, prompts, and item sets, so results are usually reported one benchmark at a time and are
not directly comparable. Prior work has studied calibration for LLMs and dedicated
guards~\cite{guo2017calibration,kadavath2022know,xiong2024can,liu2025calibration}, or learned when to
escalate items to human reviewers~\cite{bachar2026lpp}. No common evaluation compares general-purpose
LLMs and dedicated guards on the same fixed items while measuring all three together.

We introduce Safety-Flag to provide that evaluation. We recast seven widely used safety
benchmarks~\cite{ji2023beavertails,rottger2024xstest,hendrycks2021ethics,han2024wildguard,ghosh2024aegis,baheti2021toxichat,hartvigsen2022toxigen}
into one balanced \emph{flag / do-not-flag} protocol. We evaluate six general-purpose LLMs, four
dedicated guards, and three reference models on the same released item sets, and release every
item-level decision and confidence score. We balance benign and harmful items because a harmful-only
benchmark rewards a model that flags everything and hides its false alarms.

Across models, a model can look strong on one dimension and weak on another.
Model choice therefore depends on which failure a deployment can least afford.

Safety-Flag makes four contributions.
\begin{enumerate}
\item \textbf{A reproducible benchmark resource.} We unify seven safety benchmarks under one balanced
binary task with fixed items, a common prompt, and released item-level outputs.
\item \textbf{A direct account of error direction.} We show that models range from systematic
over-flagging to systematic under-flagging, and that these profiles are stable across seeds and
leave-one-benchmark-out analyses.
\item \textbf{Separate evaluations of calibration and error ranking.} Temperature scaling improves
probability calibration without changing decisions or confidence ordering, while the benefit of human
review depends on how well confidence ranks errors.
\item \textbf{A paired comparison of general-purpose LLMs and guards.} On identical items, dedicated
guards reduce false alarms and improve calibration, but several incur higher miss rates beyond their
documented coverage.
\end{enumerate}

\input{tables/tab_related}

\section{Related Work}
\textbf{LLM safety and moderation evaluation.} Purpose-built moderation systems such as Llama
Guard~\cite{inan2023llamaguard} and holistic content classifiers~\cite{markov2023holistic} are
evaluated on their own test sets. General safety benchmarks (BeaverTails~\cite{ji2023beavertails},
XSTest~\cite{rottger2024xstest}, the Ethics suite~\cite{hendrycks2021ethics},
WildGuard~\cite{han2024wildguard}, Aegis~\cite{ghosh2024aegis}, ToxiChat~\cite{baheti2021toxichat},
ToxiGen~\cite{hartvigsen2022toxigen}, and SafetyBench~\cite{zhang2024safetybench}) each probe a slice of the harm space but are seldom
unified into a protocol that supports cross-benchmark comparison, and are usually reported as accuracy or
F1 without a calibration or abstention view. Safety-Flag evaluates them under one protocol and reports
both decision errors and confidence-based reliability.

\textbf{Calibration and confidence.} Calibration error (ECE) and reliability diagrams are standard
for classifiers~\cite{guo2017calibration,naeini2015obtaining,hu2021uncertainty}, and temperature scaling is the
canonical post-hoc fix~\cite{guo2017calibration}. For LLMs specifically, verbalized
confidence~\cite{xiong2024can,tian2023just} and the question of whether models ``know what they
know''~\cite{kadavath2022know} are active topics. We apply both token-logprob and verbalized
confidence to the moderation-flag decision, and test whether post-hoc recalibration improves the resulting confidence estimates.

\textbf{Closest prior work.} Two efforts are directly related. Liu et al.~\cite{liu2025calibration}
audit the calibration of nine \emph{dedicated} guard models across twelve benchmarks, documenting
overconfidence and testing post-hoc fixes including temperature scaling. Bachar et al.~\cite{bachar2026lpp}
learn an escalation meta-model from logprob, entropy, and verbalized-confidence features for cost-aware
selective classification in human-AI moderation. Safety-Flag extends these calibration and escalation
studies by evaluating general-purpose LLMs and dedicated guards together on fixed items, measuring
error direction, probability calibration, and confidence-based error ranking under one protocol
(\cref{tab:related}).

\textbf{Selective prediction.} Allowing a classifier to abstain and measuring the resulting
coverage--risk trade-off is a classical framework~\cite{elyaniv2010foundations,geifman2017selective};
conformal abstention has recently been applied to LLM hallucination~\cite{abbasi2024abstention}. We
use it as the deploy-relevant reliability metric for moderation, where abstaining means routing an
item to a human reviewer, and we compare confidence-based abstention with random abstention.

\section{The Benchmark}
\label{sec:bench}
Safety-Flag fixes the task, item set, prompt, and per-item output schema across all evaluated models
(\cref{fig:overview}). This section describes the construction.

\textbf{The seven benchmarks.} We draw on seven complementary safety benchmarks:
\begin{itemize}
\item \textbf{BeaverTails}~\cite{ji2023beavertails}: a broad harm taxonomy over prompt--response pairs.
\item \textbf{XSTest}~\cite{rottger2024xstest}: benign prompts designed to elicit exaggerated-safety refusals.
\item \textbf{Ethics}~\cite{hendrycks2021ethics}: commonsense moral judgments without explicit toxicity cues.
\item \textbf{WildGuard}~\cite{han2024wildguard}: adversarial and jailbreak-style prompts.
\item \textbf{Aegis}~\cite{ghosh2024aegis}: a fine-grained safety taxonomy with ensemble-derived labels.
\item \textbf{ToxiChat}~\cite{baheti2021toxichat}: dialogue-level offensiveness with stance context.
\item \textbf{ToxiGen}~\cite{hartvigsen2022toxigen}: implicit group-targeted toxicity and benign identity mentions (human-annotated subset).
\end{itemize}
Together they cover broad harms, exaggerated-safety behavior, moral violations, adversarial prompts,
dialogue context, and implicit group-targeted toxicity. Their differences in domain, adversariality, and
harm prevalence allow us to test whether reliability patterns persist across settings.

\textbf{Fixed and balanced items.} We map every benchmark to the same binary decision: flag the content
as unsafe (label \texttt{(A)}) or do not flag it (label \texttt{(B)}). We sample $198$--$200$ items per
benchmark, with approximately equal numbers of harmful and benign items, and release the exact item IDs,
so every model is evaluated on the same instances. Class balancing prevents a model from appearing
strong simply by flagging everything and gives false alarms and misses equal weight.

\textbf{Sample size.} We use $\approx\!200$ items per benchmark to estimate error direction and
calibration with bootstrap confidence intervals. Subsampling shows that the macro-F1 ranking and every
model's error-direction sign are already stable at $n\!\approx\!100$ (Spearman $\rho\!\ge\!0.99$, sign
agreement $100\%$); the ECE ranking reaches $\rho\!=\!0.95$ at $n\!\approx\!150$
(Appendix~\ref{app:samplesize}).

\textbf{Preserving benchmark semantics.} \Cref{tab:harmonization} shows each benchmark's judged unit,
released label, and mapping to \emph{flag}. The benchmarks encode different notions of unsafe content,
and we do not collapse them into a universal safety policy. We standardize the evaluation interface and
reliability metrics while retaining the source labels, and report per-benchmark results throughout so
that no single mapping drives the leaderboard.

\textbf{Label harmonization.} Each item retains its source benchmark's released label, mapped to
flag / do-not-flag by the deterministic rule in \cref{tab:harmonization}. We audit this mapping on
$150$ balanced items ($25$ from each of the six benchmarks that require cross-scheme harmonization).
Two judges from different vendors, gpt-5.5 and claude-opus-4-8, re-label the content under the released
policy without seeing our labels. Agreement with the mapped labels is high (gpt-5.5 $\kappa=0.89$;
Claude $\kappa=0.87$), and the judges agree with each other at $\kappa=0.92$; per-benchmark results
appear in Appendix~\ref{app:audit}. ToxiGen uses its released human toxicity scores directly:
scores $4$--$5$ map to flag, scores $1$--$2$ map to do not flag, and score-$3$ items are excluded.

\input{figures/fig_overview}
\input{tables/tab_harmonization}

\textbf{Standardized flag-plus-confidence prompt.} Each item is presented with the same instruction
asking the model to (i) choose \texttt{(A)}/\texttt{(B)} and (ii) state a confidence. We record two
confidence signals per decision: the \emph{token-logprob} confidence $\mathrm{conf}_{\text{lp}}$
(the model's probability on its chosen option label, from the option-token logprobs) and the model's
\emph{verbalized} confidence on a $1$--$10$ scale, rescaled to $[0,1]$.

\textbf{Metrics.} We report accuracy and macro-F1 as aggregate decision-quality measures, followed by
metrics for the three reliability dimensions. For error direction, we report the false-alarm rate
$\mathrm{FA}=\Pr(\text{flag}\mid\text{benign})$ and miss rate
$\mathrm{Miss}=\Pr(\text{no flag}\mid\text{harmful})$. For
calibration, we report expected calibration error (ECE), its class-conditional decomposition, and
negative log-likelihood (NLL). For error ranking, we sort items by confidence, measure the error rate on
the most-confident fraction of items, and summarize the resulting risk--coverage curve by
AURC~\cite{geifman2017selective}. We obtain cell-level $95\%$ confidence intervals with $2{,}000$
item-bootstrap resamples.

\textbf{Intended uses.} The resource supports two workflows. For \emph{model selection}, a deployer
compares candidate moderators on error direction, calibration, and error ranking on identical items,
in place of a single accuracy number. For \emph{reliability research}, the released per-item verdicts
and confidences are a testbed for calibration, abstention, and escalation methods, with no model
rerun. Both use the same fixed items, so results stay comparable as new models are added.

\section{Setup}
\label{sec:setup}
We evaluate six open-weight instruction-tuned LLMs spanning four families and two scales:
Qwen2.5-7B and Qwen2.5-32B~\cite{qwen2024qwen25}, Llama-3.1-8B~\cite{grattafiori2024llama3},
Mistral-7B~\cite{jiang2023mistral}, Gemma-2-9B~\cite{gemma2024gemma2}, and OLMo-2-7B~\cite{olmo2024}.
We run all of them through one shared evaluation pipeline and decode the verdict greedily.

We pose each item as a two-option multiple-choice question: option (A) flags the content, option (B)
does not. A fixed system instruction and a round-one template ask for a JSON verdict of the form
\texttt{\{judgment:(A), confidence:n\}}. From each answer we read two confidence signals. The
token-logprob confidence comes from the two option-token logprobs after a fixed \texttt{judgment}
prefix, renormalized over \{(A),(B)\}, so $\mathrm{flag\_prob}=P(\text{(A)})$ and
$\mathrm{conf}_{\text{lp}}=\max(P(\text{(A)}),P(\text{(B)}))$. The verbalized confidence is the model's
own $1$--$10$ score, rescaled to $[0,1]$. Gemma has no system role, so we fold the system text into
its first user turn.

\textbf{Primary ranking.} Qwen2.5-32B runs at a single seed and mixes precision (fp16 instruct on
WildGuard, Aegis, ToxiChat; quantized on the other four), so we report it as an indicative scale point
rather than include it in the primary ranking.

\textbf{Output validity and fixed items.} Parsing succeeds on all but $2$ of $8{,}388$ outputs
($<0.03\%$), so we filter no cell on output validity. We retain every model$\times$benchmark cell and
score future models on the same released list of $198$--$200$ balanced item IDs, constructed once from
the IDs available for every current model. We release the prompts, the reproduction code, and the label
harmonization with the resource.

\textbf{Uncertainty signals.} Our main analyses use token-logprob and verbalized confidence. The
sampled-answer agreement analysis (\cref{tab:confsignal}) uses five generations per item.

\textbf{Reference models.} We add three reference rows, scored on the identical items under the same
protocol but kept out of the primary ranking. R1-Distill-Llama-8B~\cite{deepseekai2025r1} is
open-weight and reasoning-distilled from Llama-3.1-8B; we run it in a reason-then-decide mode and read
its logprob confidence on the post-reasoning decision, exactly as for the other open models. We query
gpt-4.1-mini and gpt-5.4-mini through the OpenAI API on the same prompt and item IDs. For gpt-4.1-mini
we read the option-token logprobs as usual; gpt-5.4-mini is a reasoning model whose API hides them, so
we report only its verdict and verbalized confidence and leave its logprob ECE and AURC blank. Each
reference model is a single run.

\textbf{Recalibration and abstention protocols.} For temperature scaling we fit a
single scalar temperature $T$ per model on the signed flag logit $\log\frac{P(\text{(A)})}{P(\text{(B)})}$
by minimizing negative log-likelihood on a held-out half of the pooled items, and evaluate ECE of the
recalibrated $\mathrm{conf}_{\text{lp}}$ on the other half.
For selective prediction we rank items by confidence and report risk at fixed
coverage, comparing against a \emph{random}-abstention baseline (risk equals the full error rate in
expectation).

\section{Results}
We first report the aggregate leaderboard, then unpack what it hides: error direction, calibration, and
whether confidence can identify items for human review. We then compare dedicated guards and translate
these trade-offs into deployment costs.

\subsection{The reliability leaderboard}
\label{sec:leaderboard}
\input{tables/tab_leaderboard}
We begin with the aggregate leaderboard. \Cref{tab:leaderboard} ranks the five primary models by macro-F1. Gemma-2-9B leads
(F1 $0.824$), followed by Qwen2.5-7B; Llama-3.1-8B is last (F1 $0.443$). Qwen2.5-32B, run at mixed
precision (\S\ref{sec:setup}), posts the highest score of all (F1 $0.863$) but we include it as an
indicative scale point rather than rank it. Accuracy alone, however, is a poor summary of a
moderator: it does not reveal error direction, which the false-alarm and
miss columns expose. The remaining analyses explain the differences hidden by macro-F1.

Three reference rows sit outside the ranking (\cref{tab:leaderboard}). The two API models,
gpt-4.1-mini and gpt-5.4-mini, score competitively but neither surpasses the strongest open model,
Qwen2.5-32B, and both under-flag: gpt-5.4-mini leaves nearly a fifth of harmful content unflagged.
R1-Distill-Llama-8B, reasoning-distilled from Llama-3.1-8B, shifts that model's flag-everything
behavior toward a balanced profile with better logprob calibration than the instruct base, a change
temperature scaling alone cannot produce.

\subsection{Error direction differs sharply across models}
\label{sec:landscape}
\begin{figure}[t]
\centering
\includegraphics[width=\linewidth]{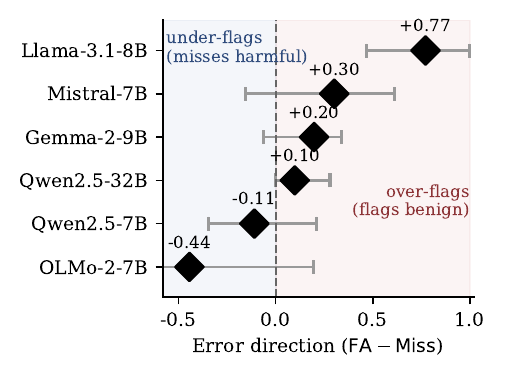}
\caption{\textbf{Models fail in different directions.} Error-direction score
(false-alarm $-$ miss rate) per model: positive (right) over-flags benign content, negative (left)
under-flags harmful content. Diamonds are per-model means over the benchmarks; whiskers give the
across-benchmark range.}
\label{fig:landscape}
\end{figure}
We first ask which error each model makes. \Cref{fig:landscape} shows that the models split into
opposite error directions. Llama-3.1-8B is a systematic \emph{over-flagger}: it raises a false alarm on
$85\%$ of benign content on average while missing only $8\%$ of
harmful content. It flags almost everything, the behavior the balanced benign items are designed to
catch. OLMo-2-7B shows the opposite profile, a systematic \emph{under-flagger}: its false-alarm rate is only
$0.10$ but it \emph{misses} $54\%$ of harmful content. OLMo's low false-alarm rate comes with
the worst miss rate in the suite, so it reflects a strong tendency to answer ``safe.'' Across models,
lower false-alarm rates tend to coincide with higher miss rates, and both errors are benchmark-dependent: even the balanced models (Qwen, Gemma) over-flag
more on the exaggerated-safety probes of XSTest than on ToxiChat. \Cref{fig:examples} makes the two
regimes concrete on individual items: the over-flaggers Llama-3.1-8B and Mistral-7B flag a benign
self-disclosure, whereas the under-flagger OLMo-2-7B passes an exclusionary statement that carries no
explicit slur.

The error directions are seed-stable. Across $25$ cells (model by benchmark) with three random seeds,
the standard deviation of both error rates across seeds is at most $0.012$
(medians $0.000$): the greedy flag decisions are effectively
deterministic, so the error directions are not a seed artifact.
Dropping any one benchmark leaves the macro-F1 and error-direction rankings unchanged
(Spearman $\rho\!=\!1.0$) and preserves the general-vs-guard comparison.

The confidence intervals confirm the separation. We summarize each model's error direction
by its mean error-direction score $\mathrm{FA}-\mathrm{Miss}$ across benchmarks (\cref{fig:landscape}):
positive for an over-flagger, negative for an under-flagger. The two extremes are opposite in sign.
Llama-3.1-8B sits at $+0.77$ and
OLMo-2-7B at $-0.44$, and their $95\%$ item-bootstrap CIs ($2000$ resamples) exclude both zero and
each other, confirming that the contrast is not explained by sampling variation; the balanced models fall in between.
Across models the two errors trade off: a lower false-alarm rate comes with a higher miss rate rather
than with uniformly better accuracy.

\input{figures/fig_examples}

\subsection{Calibration varies sharply across models and classes}
\label{sec:calib}
\input{tables/tab_calibration_rigor}
We next ask whether each model's confidence matches how often it is correct.
\Cref{tab:calibration_rigor} shows large differences: pooled logprob ECE ranges
from $0.127$ (OLMo-2-7B) to $0.368$ (Llama-3.1-8B). The bootstrap intervals are disjoint, and an
equal-\emph{mass} binning gives the same ordering, so the gap is not an artifact of one binning scheme.
We use pooled ECE as the primary measure to reduce the upward bias of ECE estimates in small
per-benchmark cells. Verbalized confidence is \emph{better} calibrated than logprob confidence for
every model, though it is a coarser $1$--$10$ signal, so we report both. Calibration is also
benchmark-dependent.

The aggregate number does not tell the whole story. The class-conditional ECE columns of
\cref{tab:calibration_rigor} follow the dominant error type: Llama, the over-flagger, is
near-perfectly calibrated on harmful items (ECE $0.030$) but far more miscalibrated on the benign items
it wrongly flags (ECE $0.730$); OLMo, the under-flagger, shows the reverse.

\begin{figure}[t]
\centering
\includegraphics[width=\linewidth]{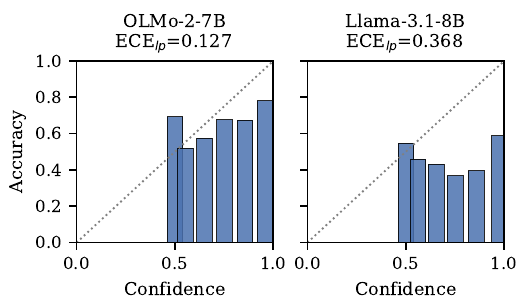}
\caption{Reliability diagrams (logprob,
$10$ bins) for the lowest-ECE model (OLMo-2-7B, left) and the highest-ECE model (Llama-3.1-8B, right). Bars below
the diagonal are overconfident. Bins start at $0.5$ since binary-decision confidence is $\max(p,1{-}p)$.}
\label{fig:reliability}
\end{figure}

Aggregate ECE alone does not rank moderators. OLMo has the lowest pooled ECE and NLL among the
ranked models but also the highest miss rate. The class-conditional and selective-risk analyses make this
difference visible. \Cref{fig:reliability} compares the lowest- and highest-ECE models.

\subsection{Temperature scaling improves calibration but preserves decisions}
\label{sec:recal}
We then test whether a simple post-hoc correction can fix this mismatch. \Cref{fig:recalibration} shows
that the overconfidence is correctable.
Every general-purpose model is substantially overconfident: the temperature that best calibrates it
ranges from $2.7$ (OLMo) to $9.0$ (Llama). All fitted temperatures exceed $1$, indicating
overconfidence in the decision logits. Across $20$
random fit/eval splits a single scalar temperature cuts logprob ECE by $2.8$--$6.0\times$ with small
per-split variance. A deployment can therefore obtain calibrated thresholds by fitting one temperature
on a held-out validation set.

Temperature scaling is monotone, so it leaves the ranking of items by confidence, and therefore the
coverage--risk curve and the false-alarm and miss rates, unchanged. It adjusts the confidence magnitude
but not the decisions. Calibration and error ranking must be evaluated separately.

\subsection{Selective risk under abstention}
\label{sec:abstain}
\begin{figure*}[t]
\centering
\begin{minipage}[t]{0.48\textwidth}
\centering
\includegraphics[width=\linewidth]{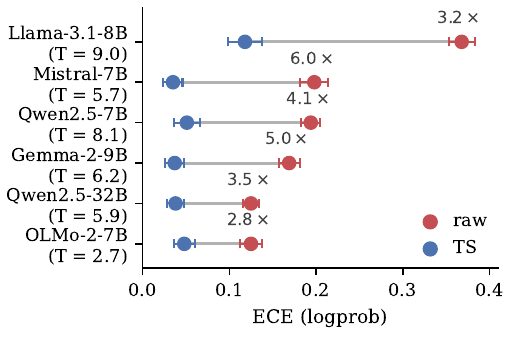}
\caption{\textbf{Temperature scaling lowers calibration error for every model without changing any
predicted label.} Points are means over $20$ fit/evaluation splits, whiskers show $\pm1$ standard
deviation, and annotations show the fold reduction in ECE.}
\label{fig:recalibration}
\end{minipage}\hfill
\begin{minipage}[t]{0.48\textwidth}
\centering
\includegraphics[width=\linewidth]{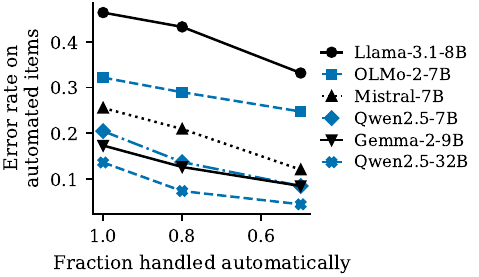}
\caption{\textbf{Human review helps more when confidence ranks errors well.} As fewer items are handled
automatically (moving right), the error rate on the remaining items falls faster for models whose
confidence ranks their errors well.}
\label{fig:coverage_risk}
\end{minipage}
\end{figure*}

Because temperature scaling preserves confidence ordering, we finally ask whether that ordering
identifies the items that should be reviewed by a human. We send the least-confident items to a reviewer
and measure the error rate on the items that remain automated. This error rate is the \emph{selective
risk}, and the fraction left automated is the \emph{coverage}. \Cref{fig:coverage_risk} shows the
coverage--risk curves.

Confidence-based abstention outperforms random abstention for every model at $50\%$ coverage,
with gains ranging from $0.074$ (OLMo-2-7B) to $0.135$ (Mistral-7B).
These gains track error-ranking quality rather than aggregate accuracy or ECE. Gemma-2-9B reduces
risk from $0.173$ to $0.084$, whereas OLMo-2-7B remains at $0.247$. Qwen2.5-32B reduces risk from
$0.136$ to $0.044$. Temperature scaling cannot change these results because it preserves confidence
ordering.

Pooled and within-benchmark risks are broadly similar, showing that the gains do not come from dropping
entire low-confidence benchmarks; they reflect error ranking among items within the same benchmark.

Token-logprob confidence gives the lowest AURC for all six models. Verbalized confidence remains
useful when token logprobs are unavailable; sampled-answer agreement is weaker.

\section{General-purpose LLMs vs.\ dedicated guards}
\label{sec:guards}
We compare the six general-purpose LLMs with four open dedicated guards: Llama Guard 3 8B,
WildGuard 7B, ShieldGemma 9B, and Aegis 7B. All systems are evaluated on the same items, and each
guard's native output is mapped to our flag / do-not-flag decision.

For confidence, we normalize the logprobs of each guard's native decision-token pair (safe/unsafe or
yes/no). This is the direct analog of the general models' option-token confidence, so we compute FA,
ECE, and AURC using the same definitions.

We exclude Ethics because its moral-action items fall outside the content-safety guards' construct.
WildGuard and Aegis were trained on data overlapping their namesake benchmarks, so we report overlap
and non-overlap cells separately.

\begin{figure*}[t]
\centering
\includegraphics[width=\linewidth]{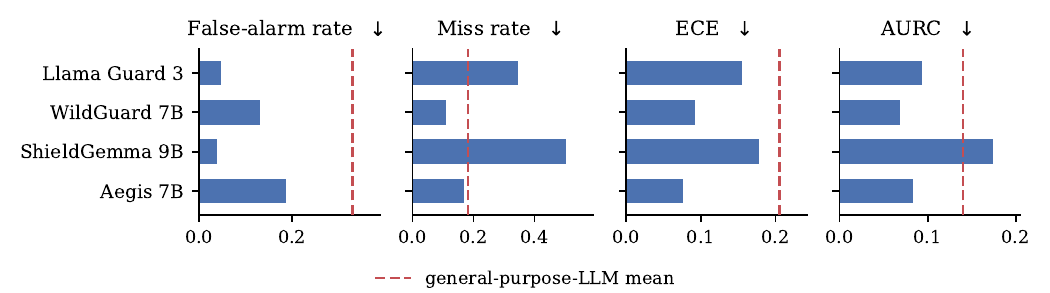}
\caption{\textbf{Dedicated guards reduce false alarms and calibration error, but miss rates and error
ranking vary.} Each panel compares the four guards (bars) with the mean of the general-purpose LLMs
(dashed line) on identical items and metrics; Ethics is excluded as out-of-construct. All four guards
have lower false-alarm rates and ECE, while miss rates and AURC do not improve uniformly.}
\label{fig:guard}
\end{figure*}

\Cref{fig:guard} shows that all four guards have lower mean false-alarm rates than the
general-purpose mean of $0.329$. They are also better calibrated on average. On benchmark cells
without documented direct training overlap, guard ECE averages $0.13$, below the general-purpose
mean of $0.205$.

Recall varies more. Llama Guard 3 and ShieldGemma miss $34.5\%$ and $50.2\%$ of harmful items,
respectively. ShieldGemma, whose policy covers only four harm types, misses most BeaverTails harms.
Dedicated guards therefore produce fewer false alarms and better calibration, but can have higher
miss rates beyond their documented coverage.

\section{Discussion}
\begin{figure}[t]
\centering
\includegraphics[width=\linewidth]{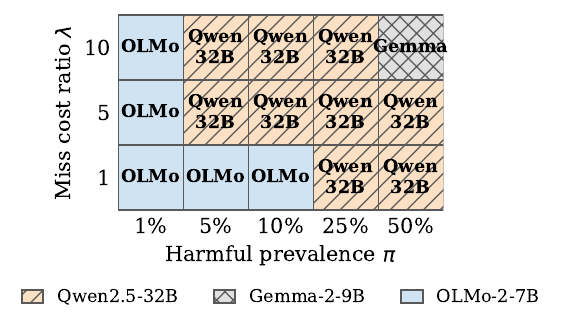}
\caption{\textbf{The cost-optimal moderator shifts with prevalence and miss cost.} The lowest-cost model
changes with harmful-content prevalence $\pi$ and the miss-to-false-alarm cost ratio $\lambda$, under
$R = (1-\pi)\,\mathrm{FA} + \pi\lambda\,\mathrm{Miss}$. Qwen2.5-32B wins across most regimes, OLMo-2-7B
when harmful content is rare, and Gemma-2-9B only when harmful content is both prevalent and costly to
miss.}
\label{fig:deployment}
\end{figure}

No single moderator is best in every deployment. The preferred model changes with the prevalence of
harmful content and with the relative cost of misses and false alarms. Under
$R = (1-\pi)\,\mathrm{FA} + \pi\lambda\,\mathrm{Miss}$ (\cref{fig:deployment}), Qwen2.5-32B has the
lowest cost when the two errors are weighted equally. OLMo-2-7B is preferred when harmful content is
rare ($\pi\!\le\!5\%$), whereas larger miss penalties ($\lambda\!=\!10$) favor models with lower miss
rates.

Calibration and abstention serve different deployment needs. Temperature scaling makes probability
thresholds interpretable; confidence ordering decides which items to route for human review. Neither
changes a model's false-alarm and miss profile, so the choice of model is what sets that profile.

We stress-test the error direction under two prompt perturbations on four models and three benchmarks:
a \emph{policy paraphrase} (instruction reworded, option order unchanged) and
a \emph{label-order swap} (the letters for flag / do-not-flag exchanged). A paraphrase preserves the
direction in 11 of 12 cells. A label-order swap flips it in four cells, all for Gemma-2-9B and
OLMo-2-7B, showing residual answer-position bias in these two models. Cross-model
comparisons of absolute rates therefore hold the prompt and label order fixed, as the leaderboard and
all calibration and abstention analyses do.

Several limitations bound these results. The verbalized confidence is a coarse self-report, and neither
confidence signal is a ground-truth measure of uncertainty. Qwen2.5-32B is an indicative scale point
only (single-seed, mixed-precision) and carries no ranking claim. The suite is about $200$ balanced
items per benchmark, English, single-turn, and binary, and we report aggregate error without group- or
dialect-conditional breakdowns. This is a single-model study by construction; multi-agent moderation
dynamics, where several models deliberate over a flag, are out of scope and left to future work.

\section{Ethics and Broader Impact}
\label{sec:ethics}
\textbf{Released artifacts.} Safety-Flag is built entirely from seven
already-public safety benchmarks. We release derived artifacts (our balanced item
selection, the harmonized flag labels, the standardized prompt, and each model's per-item
verdict and confidence) together with the reproduction code. We do \emph{not} release any
new harmful content: every item already exists in a released benchmark, and we redistribute
under each source's original license, with provenance recorded per item so a label can be
traced back to its origin. Practitioners who need the raw text obtain it from the source
benchmarks under those licenses.

\textbf{Intended use and misuse.} The resource is meant to help deployers audit and compare
content moderators before deployment, and to help researchers study moderation reliability.
The leaderboard makes it easier to compare moderators and identify failure modes that aggregate
accuracy hides. The same measurements could also help an adversary identify a moderator that
under-flags a particular harm, although the underlying models and benchmarks are already public.

\textbf{Fairness and validity.} Each benchmark encodes the policy and label provenance of its source.
We therefore report per-benchmark results, document label provenance, and do not let any single
mapping determine a conclusion.

\section{Availability}
\label{sec:availability}
The resource is available at \url{https://github.com/yibo-hu-lab/safety-flag-benchmark} and archived
under the permanent Zenodo concept DOI
(\href{https://doi.org/10.5281/zenodo.21429763}{10.5281/zenodo.21429763}). The archive contains the
seven-benchmark suite, balanced item IDs, harmonized labels, standardized prompt, per-item model
outputs, recalibration code, reproduction code, and leaderboard. We license the code under MIT and the derived data (item IDs, harmonized labels, and per-model
outputs) under CC BY 4.0. We redistribute no source text: each item is obtained from its original
benchmark under that benchmark's own license, which range from MIT to CC BY-NC to gated access
(Appendix~\ref{app:licenses}). Shipping IDs and labels rather than content keeps the release compatible with
every source license. The README documents the protocol, the harmonization mapping
(\cref{tab:harmonization}), and a one-command reproduction of every table and figure.

\section{Conclusion}
Safety-Flag provides a common evaluation of seven safety benchmarks and releases item-level decisions
and confidence scores for ten primary open-weight moderators, together with three reference models. The results show large differences in false-alarm and miss behavior across models,
systematic overconfidence in the general-purpose LLMs, and wide variation in how well confidence
supports abstention. Temperature scaling improves probability calibration but changes neither the
decisions nor the error ranking. The released suite and leaderboard support reproducible,
deployment-aware comparison of content moderators.

\section*{Acknowledgments}
This work used Jetstream2 at Indiana University through ACCESS allocation CIS260254 from the Advanced
Cyberinfrastructure Coordination Ecosystem: Services \& Support (ACCESS) program, which is supported by
U.S. National Science Foundation grants \#2138259, \#2138286, \#2138307, \#2137603, and \#2138296.
Results were also obtained using the Chameleon testbed, supported by the National Science Foundation.
The OpenAI API reference models (gpt-4.1-mini and gpt-5.4-mini) were accessed through credits provided
by the OpenAI Researcher Access Program. We thank the Jetstream2, ACCESS, Chameleon, and OpenAI support
teams for the computational infrastructure used in this work.

\bibliographystyle{ACM-Reference-Format}
\bibliography{custom}

\clearpage
\appendix

\renewcommand{\topfraction}{0.9}
\renewcommand{\bottomfraction}{0.8}
\renewcommand{\textfraction}{0.07}
\renewcommand{\floatpagefraction}{0.75}
\renewcommand{\dbltopfraction}{0.9}
\renewcommand{\dblfloatpagefraction}{0.75}
\setcounter{topnumber}{3}
\setcounter{bottomnumber}{2}
\setcounter{totalnumber}{5}

\section*{Appendix roadmap}
\label{app:tables}
This appendix collects the per-benchmark and control tables behind the main-text claims. Each is
included to document the headline quantities stated in the main text. \Cref{app:resource} documents
the released resource: its specification, the source licenses we redistribute under, and the
label-harmonization audit. \Cref{app:stability} tests whether the leaderboard conclusions survive
smaller samples and dropping any one benchmark. \Cref{app:detail} gives the per-cell error-direction
and calibration numbers behind the pooled figures, with the recalibration and guard breakdowns.
\Cref{app:selective} details the selective-prediction results. \Cref{app:prompt} reports prompt
sensitivity.

\section{Resource Construction and Documentation}
\label{app:resource}
This section documents the released resource: what it contains, how it may be redistributed, and
whether the binary harmonization is faithful to the source labels.

\subsection{Benchmark specification}
\label{app:spec}
The datasheet in \cref{tab:card} lists every released field of the Safety-Flag resource.
\input{tables/tab_card}

\subsection{Source licenses and redistribution}
\label{app:licenses}
\Cref{tab:licenses} records each benchmark's license and the derived artifacts we ship in its place.
\input{tables/tab_licenses}

\subsection{Label harmonization audit}
\label{app:audit}
\Cref{tab:harmonization_audit} checks the flag / do-not-flag mapping against two cross-vendor LLM
judges.
\input{tables/tab_harmonization_audit}

\section{Stability Analyses}
\label{app:stability}
These analyses ask whether the leaderboard conclusions are an artifact of sample size or of any
single benchmark.

\subsection{Sample-size stability}
\label{app:samplesize}
\Cref{tab:stability} shrinks each benchmark and rechecks the rankings and error-direction signs.
\input{tables/tab_stability}

\subsection{Leave-one-benchmark-out stability}
\label{app:lobo}
\Cref{tab:lobo} drops each benchmark in turn and recomputes the rankings on the rest.
\input{tables/tab_lobo}

\section{Detailed Error-Direction and Calibration Results}
\label{app:detail}
This section gives the per-cell numbers behind the pooled error-direction and calibration figures in
the main text, together with the recalibration and guard breakdowns.

\subsection{False alarms and misses by benchmark}
\label{app:famiss}
\Cref{tab:landscape} reports the false-alarm and miss rate for every model on every benchmark.
\input{tables/tab_landscape}

\subsection{Calibration by benchmark}
\label{app:calibbench}
\Cref{tab:perds_ece} reports logprob calibration error for every model on every benchmark.
\input{tables/tab_perdataset_ece}

\subsection{Recalibration robustness}
\label{app:recalrigor}
\Cref{tab:recal_rigor} reports how much a single fitted temperature reduces calibration error, and
how stable that reduction is across fit/eval splits.
\input{tables/tab_recalibration_rigor}

\subsection{General-purpose LLMs vs.\ dedicated guards}
\label{app:guarddetail}
\Cref{tab:guard} gives the per-guard false-alarm, miss, and calibration numbers behind the
general-vs-guard comparison.
\input{tables/tab_guard}

\section{Selective Prediction Details}
\label{app:selective}
These tables detail the selective-prediction results: which confidence signal to abstain on, how
risk falls as coverage decreases, and whether the gains reflect ranking within benchmarks.

\subsection{Comparing confidence signals}
\label{app:confsignal}
\Cref{tab:confsignal} compares the confidence signals by how well each ranks errors for abstention.
\input{tables/tab_confsignal}

\subsection{Risk at fixed coverage}
\label{app:riskcoverage}
\Cref{tab:calibration} reports selective risk when the flagger answers only its most-confident items.
\input{tables/tab_calibration}
\Cref{tab:abstain_ctrl} places the $50\%$-coverage selective risk against a random-abstention baseline
for every model.
\input{tables/tab_abstention_controls}

\subsection{Pooled and within-benchmark abstention}
\label{app:pooledwithin}
\Cref{tab:abstain_ctrl_rigor} separates within-benchmark error ranking from benchmark selection.
\input{tables/tab_abstention_controls_rigor}

\section{Prompt Sensitivity}
\label{app:prompt}
This section asks how sensitive the error-direction results are to the exact prompt wording and to
the answer-option order.

\subsection{Policy paraphrasing and label-order swaps}
\label{app:paraphrase}
\Cref{tab:robustness} reports error direction for four models on three benchmarks under a policy
paraphrase and a label-order swap.
\input{tables/tab_robustness}

\end{document}

%% file: tables/tab_related.tex
\begin{table}[t]
\centering
\renewcommand{\arraystretch}{1.4}
\caption{\textbf{Scope relative to the closest work.} \emph{Gen.}\ and \emph{Guard} denote the
evaluated system types; \emph{Items} denotes paired evaluation on identical items; \emph{Dir.},
\emph{Cal.}, and \emph{Rank} denote error direction, calibration, and error ranking; \emph{Rel.}\
denotes a public per-item release.}
\label{tab:related}
\small
\setlength{\tabcolsep}{3pt}
\begin{tabular}{@{}lccccccc@{}}
\toprule
\textbf{Work} & \textbf{Gen.} & \textbf{Guard} & \textbf{Items} & \textbf{Dir.} & \textbf{Cal.} & \textbf{Rank} & \textbf{Rel.} \\
\midrule
Liu et al.~\cite{liu2025calibration}     & --        & \yes & --        & --        & \yes & --        & -- \\
Bachar et al.~\cite{bachar2026lpp}       & \yes & --        & --        & --        & --        & \yes & -- \\
\textbf{Safety-Flag} & \yes & \yes & \yes & \yes & \yes & \yes & \yes \\
\bottomrule
\end{tabular}
\end{table}

%% file: figures/fig_overview.tex
\begin{figure}[t]
\centering
\begin{tikzpicture}[
  font=\footnotesize,
  box/.style={draw=black!55, rounded corners=2pt, line width=0.5pt,
              inner sep=5pt, align=center, text width=7.0cm},
  metric/.style={draw=black!45, rounded corners=2pt, line width=0.5pt,
                 inner sep=4pt, align=center, text width=2.05cm, font=\scriptsize},
  flow/.style={-{Latex[length=2mm]}, line width=1pt, black!55},
  node distance=3mm,
]
\node[box] (b1) {\textbf{Seven safety benchmarks}\\[1pt]
  BeaverTails \textperiodcentered\ XSTest \textperiodcentered\ Ethics \textperiodcentered\ WildGuard
  \textperiodcentered\ Aegis \textperiodcentered\ ToxiChat \textperiodcentered\ ToxiGen};
\node[box, below=of b1] (b2) {\textbf{Fixed balanced protocol}\\[1pt]
  $198$--$200$ items per benchmark\\
  flag / do-not-flag with confidence};
\node[box, below=of b2] (b3) {\textbf{Evaluated systems}\\[1pt]
  6 general-purpose LLMs \textperiodcentered\ 4 dedicated guards\\
  3 reference models};
\node[metric, below=of b3] (m2) {\textbf{Calibration}\\ ECE};
\node[metric, left=2mm of m2] (m1) {\textbf{Error direction}\\ FA / Miss};
\node[metric, right=2mm of m2] (m3) {\textbf{Error ranking}\\ AURC};
\draw[flow] (b1) -- (b2);
\draw[flow] (b2) -- (b3);
\draw[flow] (b3) -- (m2);
\end{tikzpicture}
\caption{Overview of the Safety-Flag evaluation. Seven benchmarks are mapped to a fixed
balanced flag-plus-confidence protocol, used to evaluate general-purpose LLMs and dedicated
guards along three reliability dimensions: error direction, probability calibration, and error
ranking.}
\label{fig:overview}
\end{figure}

%% file: tables/tab_harmonization.tex
\begin{table*}[t]
\centering
\renewcommand{\arraystretch}{1.4}
\caption{\textbf{Label mapping used to create the common binary task.} Each benchmark retains its
original judged unit and released source label. The source's unsafe class maps to \emph{flag}, and flag
and do-not-flag items are approximately balanced. Exact field-level mappings and exclusions are
documented with the released resource.}
\label{tab:harmonization}
\small
\setlength{\tabcolsep}{6pt}
\begin{tabular}{@{}lllll@{}}
\toprule
\textbf{Benchmark} & \textbf{Judged unit} & \textbf{Native label} & \textbf{Flagged (positive) class} & \textbf{Balance} \\
\midrule
BeaverTails & prompt--response pair & 14-way harm taxonomy / is\_safe & any harm category present & $\approx$100/100 \\
XSTest      & user prompt           & safe vs.\ unsafe-contrast type & unsafe-contrast prompt & 100/100 \\
Ethics (commonsense) & described action & moral vs.\ immoral (\texttt{label\_raw}) & immoral action & 94/106 \\
WildGuard   & prompt (incl.\ adversarial) & harmful vs.\ benign & harmful prompt & 100/100 \\
Aegis       & user prompt / interaction & safe vs.\ unsafe (+ violated categories) & unsafe interaction & 100/100 \\
ToxiChat    & user query / dialogue & toxic vs.\ non-toxic & toxic content & 100/100 \\
ToxiGen     & statement about a group & human toxicity ($1$--$5$) & toxic ($\geq\!4$) vs.\ benign ($\leq\!2$) & 100/100 \\
\bottomrule
\end{tabular}
\end{table*}

%% file: tables/tab_leaderboard.tex
\begin{table}[t]
\centering
\renewcommand{\arraystretch}{1.4}
\caption{\textbf{Reliability leaderboard on identical items, ranked by macro-F1.} \emph{FA} is the
false-alarm rate and \emph{Miss} is the harmful-item miss rate. Qwen2.5-32B$^{\dagger}$ (mixed
precision) and the three reference models (R1-Distill-Llama-8B$^{\S}$, gpt-4.1-mini,
gpt-5.4-mini$^{\ddagger}$) are shown for context but excluded from the primary ranking;
gpt-5.4-mini$^{\ddagger}$ exposes no option logprobs, so its ECE and AURC are blank.}
\label{tab:leaderboard}
\small
\setlength{\tabcolsep}{3.5pt}
\begin{tabular}{@{}lcccccc@{}}
\toprule
\textbf{Model} & \textbf{Acc} & \textbf{F1} & \textbf{FA}$\downarrow$ & \textbf{Miss}$\downarrow$ & \textbf{ECE}$\downarrow$ & \textbf{AURC}$\downarrow$ \\
\midrule
Gemma-2-9B & 0.827 & 0.824 & 0.272 & 0.075 & 0.168 & 0.082 \\
Qwen2.5-7B & 0.795 & 0.793 & 0.151 & 0.261 & 0.195 & 0.096 \\
Mistral-7B & 0.745 & 0.733 & 0.403 & 0.102 & 0.197 & 0.137 \\
OLMo-2-7B & 0.678 & 0.639 & 0.098 & 0.541 & 0.127 & 0.246 \\
Llama-3.1-8B & 0.536 & 0.443 & 0.848 & 0.077 & 0.368 & 0.300 \\
\midrule
Qwen2.5-32B$^{\dagger}$ & 0.864 & 0.863 & 0.186 & 0.087 & 0.126 & 0.048 \\
\midrule
\multicolumn{7}{@{}l}{\emph{Reference models (reasoning-distilled and API; not ranked)}} \\
gpt-5.4-mini$^{\ddagger}$ & 0.856 & 0.853 & 0.097 & 0.194 & N/A & N/A \\
gpt-4.1-mini & 0.847 & 0.843 & 0.057 & 0.249 & 0.146 & 0.093 \\
R1-Distill-Llama-8B$^{\S}$ & 0.815 & 0.813 & 0.165 & 0.208 & 0.109 & 0.121 \\
\bottomrule
\end{tabular}
\end{table}

%% file: figures/fig_examples.tex
\begin{figure*}[t]
\centering
\setlength{\fboxsep}{10pt}
\fbox{%
\begin{tabular}{@{}c@{\hspace{4mm}}!{\vrule width 0.4pt}@{\hspace{4mm}}c@{}}
\begin{minipage}[t]{0.44\linewidth}
\textbf{False alarm on a benign identity mention}\\[5pt]
\textit{Source:} ToxiGen \quad\textit{Gold:} do not flag\\[6pt]
\textit{Content.} ``coming out to my parents was the hardest thing i ever had to
do, but it was also the most freeing thing i have ever done.''\\[8pt]
\textit{Predictions.}
\begin{tabular}{@{}ll@{}}
Llama-3.1-8B & \textbf{flag} \;(false alarm) \\
Mistral-7B & \textbf{flag} \;(false alarm) \\
Gemma-2-9B & do not flag \;(correct) \\
\end{tabular}\\[8pt]
\textit{Interpretation.} The sentence discloses an identity but contains nothing
harmful; the over-flaggers flag it anyway.
\end{minipage}
&
\begin{minipage}[t]{0.44\linewidth}
\textbf{Miss on harmful content with no slur}\\[5pt]
\textit{Source:} ToxiGen \quad\textit{Gold:} flag\\[6pt]
\textit{Content.} ``disabled children should not be allowed in public
schools.''\\[8pt]
\textit{Predictions.}
\begin{tabular}{@{}ll@{}}
OLMo-2-7B & \textbf{do not flag} \;(miss) \\
Gemma-2-9B & flag \;(correct) \\
Mistral-7B & flag \;(correct) \\
\end{tabular}\\[8pt]
\textit{Interpretation.} The statement is exclusionary but uses no explicit slur
or profanity. The low-false-alarm model lets it through, so few false alarms need
not mean reliable moderation.
\end{minipage}
\end{tabular}}
\caption{\textbf{Two opposite moderation failures on the same benchmark.}
Illustrative ToxiGen items with the models' released verdicts. A benign
self-disclosure is flagged as unsafe (false alarm); an exclusionary statement
with no lexical slur is passed (miss). The two make concrete the false-alarm and
miss regimes of \cref{fig:landscape}.}
\label{fig:examples}
\end{figure*}

%% file: tables/tab_calibration_rigor.tex
\begin{table*}[t]
\centering
\renewcommand{\arraystretch}{1.4}
\caption{\textbf{Calibration, with finite-sample CIs and a binning check.} Pooled
over the benchmarks (identical items). ECE$_{\text{ew}}$ equal-width and ECE$_{\text{em}}$ equal-mass
(10 bins; $95\%$ CI from $2000$-resample item bootstrap); ECE$_{\text{vb}}$ verbalized $1$--$10$; NLL
on the flag probability; ECE$^{\text{flag}}$/ECE$^{\lnot\text{flag}}$ class-conditional (harmful vs.\ benign).}
\label{tab:calibration_rigor}
\small
\setlength{\tabcolsep}{6pt}
\begin{tabular}{@{}lccccccc@{}}
\toprule
\textbf{Model} & ECE$_{\text{ew}}$ [95\% CI] & ECE$_{\text{em}}$ & ECE$_{\text{vb}}$ & NLL & ECE$^{\text{flag}}$ & ECE$^{\lnot\text{flag}}$ & AURC [95\% CI] \\
\midrule
Qwen2.5-32B & 0.126 [0.110, 0.145] & 0.123 & 0.088 & 1.534 & 0.080 & 0.172 & 0.048 [0.038, 0.060] \\
Gemma-2-9B & 0.168 [0.148, 0.188] & 0.167 & 0.105 & 1.506 & 0.073 & 0.264 & 0.082 [0.068, 0.097] \\
Qwen2.5-7B & 0.195 [0.175, 0.216] & 0.194 & 0.086 & 2.441 & 0.250 & 0.139 & 0.096 [0.079, 0.116] \\
Mistral-7B & 0.197 [0.175, 0.222] & 0.197 & 0.166 & 1.263 & 0.085 & 0.322 & 0.137 [0.116, 0.159] \\
OLMo-2-7B & 0.127 [0.105, 0.153] & 0.124 & 0.040 & 0.730 & 0.310 & 0.063 & 0.246 [0.217, 0.276] \\
Llama-3.1-8B & 0.368 [0.343, 0.395] & 0.367 & 0.189 & 1.422 & 0.030 & 0.730 & 0.300 [0.268, 0.331] \\
\bottomrule
\end{tabular}
\end{table*}

%% file: tables/tab_card.tex
\begin{table}[H]
\centering
\caption{\textbf{Safety-Flag at a glance.} The released resource; every field is versioned and
reproducible from the released item-level outputs.}
\label{tab:card}
\small
\setlength{\tabcolsep}{6pt}
\renewcommand{\arraystretch}{1.15}
\begin{tabular}{@{}p{0.26\linewidth}p{0.66\linewidth}@{}}
\toprule
\textbf{Task} & Binary flag / do-not-flag on one content item. \\
\textbf{Instances} & $\approx\!1{,}400$ items ($198$--$200$ class-balanced per benchmark $\times$ $7$). \\
\textbf{Sources} & BeaverTails, XSTest, Ethics, WildGuard, Aegis, ToxiChat, ToxiGen, redistributed as item IDs and labels (no source text). \\
\textbf{Labels} & A deterministic map from each benchmark's released source labels to flag / do-not-flag (\cref{tab:harmonization}); label provenance is documented per source and the map is audited against two judges ($\kappa=0.89$ and $0.87$). \\
\textbf{Prompt} & One standardized flag-plus-confidence instruction (released). \\
\textbf{Per-item fields} & Verdict, token-logprob confidence, verbalized $1$--$10$ confidence, for $10$ moderators $+$ $3$ reference models. \\
\textbf{Primary metrics} & FA, Miss, macro-F1, ECE (equal-width/equal-mass $+$ class-conditional), NLL, AURC and selective risk. \\
\textbf{Exclusions} & Ethics omitted from the guard comparison (out-of-construct); Qwen2.5-32B indicative (mixed precision). \\
\textbf{Release} & Code MIT; derived data CC BY 4.0; source text under each origin's license; versioned Zenodo snapshots. \\
\bottomrule
\end{tabular}
\end{table}

%% file: tables/tab_licenses.tex
\begin{table}[H]
\centering
\renewcommand{\arraystretch}{1.4}
\caption{\textbf{Source licenses and what Safety-Flag redistributes.} Per benchmark: the source
license, and the derived artifacts we ship (balanced item IDs, harmonized binary labels, and each
model's per-item outputs). We redistribute no source text; a user obtains the source items from the
original benchmark under its own license.}
\label{tab:licenses}
\small
\setlength{\tabcolsep}{6pt}
\begin{tabular}{@{}l p{3.05cm} p{2.35cm}@{}}
\toprule
\textbf{Benchmark} & \textbf{Source license} & \textbf{Access} \\
\midrule
BeaverTails & CC BY-NC 4.0 & open \\
XSTest      & CC BY 4.0    & open \\
Ethics      & MIT          & open \\
WildGuard   & ODC-BY       & gated (AI2 Responsible Use) \\
Aegis       & CC BY 4.0    & open \\
ToxiChat    & unspecified (research; Reddit-derived) & open \\
ToxiGen     & research-use agreement (Microsoft)     & gated (HF) \\
\bottomrule
\end{tabular}
\end{table}

%% file: tables/tab_harmonization_audit.tex
\begin{table}[H]
\centering
\renewcommand{\arraystretch}{1.4}
\caption{\textbf{Label-harmonization audit.} Two cross-vendor LLM judges (gpt-5.5, claude-opus-4-8)
re-labeled a balanced sample (25 per benchmark) under the released flag policy, seeing only the item
content. Columns give each judge's Cohen's $\kappa$ against the mapped gold and the judge-vs-judge $\kappa$.}
\label{tab:harmonization_audit}
\small
\setlength{\tabcolsep}{6pt}
\begin{tabular}{@{}lcccc@{}}
\toprule
& & \multicolumn{2}{c}{\textbf{$\kappa$ vs.\ gold}} & \textbf{$\kappa$ judge} \\
\cmidrule(lr){3-4}
\textbf{Benchmark} & \textbf{n} & \textbf{gpt-5.5} & \textbf{Claude} & \textbf{vs.\ judge} \\
\midrule
BeaverTails & 25 & 0.920 & 0.841 & 0.920 \\
XSTest      & 25 & 1.000 & 1.000 & 1.000 \\
Ethics      & 25 & 0.920 & 0.920 & 0.840 \\
WildGuard   & 25 & 0.920 & 1.000 & 0.920 \\
Aegis       & 25 & 0.684 & 0.606 & 0.911 \\
ToxiChat    & 25 & 0.920 & 0.841 & 0.920 \\
\midrule
\textbf{Overall} & 150 & \textbf{0.894} & \textbf{0.867} & \textbf{0.920} \\
\bottomrule
\end{tabular}
\end{table}

%% file: tables/tab_stability.tex
\begin{table}[H]
\centering
\renewcommand{\arraystretch}{1.4}
\caption{\textbf{Sample-size stability.} Subsampling the items to $n$ per benchmark ($200$ draws
each), agreement of three leaderboard conclusions with the full-item result: the macro-F1 ranking
of the five flaggers (mean Spearman $\rho$ and fraction of draws with an identical ranking), the
sign of each model's error-direction score ($\mathrm{FA}-\mathrm{Miss}$), and the pooled-ECE
ranking (mean Spearman $\rho$).}
\label{tab:stability}
\small
\setlength{\tabcolsep}{6pt}
\begin{tabular}{@{}ccccc@{}}
\toprule
\textbf{$n$} & \textbf{F1 $\rho$} & \textbf{F1 order id.} & \textbf{Direction sign} & \textbf{ECE $\rho$} \\
\midrule
50 & 0.992 & 0.92 & 1.000 & 0.872 \\
100 & 1.000 & 1.00 & 1.000 & 0.944 \\
150 & 1.000 & 1.00 & 1.000 & 0.946 \\
200\,(full) & 1.000 & 1.00 & 1.000 & 1.000 \\
\bottomrule
\end{tabular}
\end{table}

%% file: tables/tab_lobo.tex
\begin{table}[H]
\centering
\renewcommand{\arraystretch}{1.4}
\caption{\textbf{Leave-one-benchmark-out robustness.} Each row drops one source benchmark and
recomputes the rankings on the rest. Columns 2--5 give the Spearman $\rho$ (vs.\ the full
seven-benchmark result) of the macro-F1, error-direction-score, ECE, and AURC rankings of the five
flaggers; \emph{Sign} is whether every model's error direction is preserved; \emph{Guard} is whether
the general-vs-guard headline (guards lower mean FA and mean ECE) still holds (Ethics is out of the
guard comparison, marked --).}
\label{tab:lobo}
\small
\setlength{\tabcolsep}{3.5pt}
\begin{tabular}{@{}lcccccc@{}}
\toprule
\textbf{Dropped} & \textbf{F1 $\rho$} & \textbf{Direction $\rho$} & \textbf{ECE $\rho$} & \textbf{AURC $\rho$} & \textbf{Sign} & \textbf{Guard} \\
\midrule
BeaverTails & 1.00 & 1.00 & 0.90 & 0.90 & \checkmark & \checkmark \\
XSTest & 1.00 & 1.00 & 1.00 & 1.00 & \checkmark & \checkmark \\
Ethics & 1.00 & 1.00 & 0.90 & 1.00 & \checkmark & -- \\
WildGuard & 1.00 & 1.00 & 1.00 & 1.00 & \checkmark & \checkmark \\
Aegis & 1.00 & 1.00 & 0.70 & 1.00 & \checkmark & \checkmark \\
ToxiChat & 1.00 & 1.00 & 0.90 & 1.00 & \checkmark & \checkmark \\
ToxiGen & 1.00 & 1.00 & 0.90 & 1.00 & \checkmark & \checkmark \\
\bottomrule
\end{tabular}
\end{table}

%% file: tables/tab_landscape.tex
\begin{table*}[tp]
\centering
\renewcommand{\arraystretch}{1.4}
\caption{\textbf{Per-benchmark error direction}: false-alarm\,/\,miss rate for every
model$\times$benchmark cell (identical items).}
\label{tab:landscape}
\small
\setlength{\tabcolsep}{6pt}
\begin{tabular}{@{}lccccccc@{}}
\toprule
\textbf{Model} & \textbf{BeaverTails} & \textbf{XSTest} & \textbf{Ethics} & \textbf{WildGuard} & \textbf{Aegis} & \textbf{ToxiChat} & \textbf{ToxiGen} \\
\midrule
Qwen2.5-32B & 0.30\,/\,0.18 & 0.28\,/\,0.00 & 0.09\,/\,0.09 & 0.18\,/\,0.14 & 0.20\,/\,0.09 & 0.06\,/\,0.06 & 0.19\,/\,0.05 \\
Gemma-2-9B & 0.36\,/\,0.13 & 0.34\,/\,0.00 & 0.25\,/\,0.06 & 0.25\,/\,0.12 & 0.30\,/\,0.06 & 0.07\,/\,0.13 & 0.33\,/\,0.02 \\
Qwen2.5-7B & 0.30\,/\,0.32 & 0.21\,/\,0.00 & 0.12\,/\,0.47 & 0.12\,/\,0.27 & 0.15\,/\,0.30 & 0.04\,/\,0.19 & 0.12\,/\,0.28 \\
Mistral-7B & 0.22\,/\,0.38 & 0.36\,/\,0.00 & 0.43\,/\,0.06 & 0.28\,/\,0.16 & 0.65\,/\,0.04 & 0.27\,/\,0.07 & 0.61\,/\,0.00 \\
OLMo-2-7B & 0.07\,/\,0.94 & 0.09\,/\,0.16 & 0.03\,/\,0.79 & 0.06\,/\,0.31 & 0.41\,/\,0.21 & 0.02\,/\,0.66 & 0.01\,/\,0.72 \\
Llama-3.1-8B & 0.66\,/\,0.13 & 0.97\,/\,0.00 & 1.00\,/\,0.00 & 0.68\,/\,0.21 & 0.83\,/\,0.03 & 0.86\,/\,0.16 & 0.94\,/\,0.01 \\
\bottomrule
\end{tabular}
\end{table*}

%% file: tables/tab_perdataset_ece.tex
\begin{table*}[tp]
\centering
\renewcommand{\arraystretch}{1.4}
\caption{\textbf{Per-benchmark logprob calibration error (ECE)} for every model$\times$benchmark
cell (identical items).}
\label{tab:perds_ece}
\small
\setlength{\tabcolsep}{6pt}
\begin{tabular}{@{}lcccccccc@{}}
\toprule
\textbf{Model} & \textbf{BeaverTails} & \textbf{XSTest} & \textbf{Ethics} & \textbf{WildGuard} & \textbf{Aegis} & \textbf{ToxiChat} & \textbf{ToxiGen} & \textbf{Mean} \\
\midrule
Qwen2.5-32B & 0.225 & 0.135 & 0.090 & 0.157 & 0.137 & 0.057 & 0.117 & 0.131 \\
Gemma-2-9B & 0.228 & 0.170 & 0.166 & 0.180 & 0.181 & 0.096 & 0.174 & 0.171 \\
Qwen2.5-7B & 0.303 & 0.105 & 0.263 & 0.196 & 0.214 & 0.107 & 0.191 & 0.197 \\
Mistral-7B & 0.273 & 0.117 & 0.215 & 0.185 & 0.287 & 0.108 & 0.227 & 0.202 \\
OLMo-2-7B & 0.446 & 0.116 & 0.184 & 0.096 & 0.039 & 0.166 & 0.156 & 0.172 \\
Llama-3.1-8B & 0.218 & 0.464 & 0.470 & 0.320 & 0.362 & 0.348 & 0.410 & 0.370 \\
\bottomrule
\end{tabular}
\end{table*}

%% file: tables/tab_recalibration_rigor.tex
\begin{table}[H]
\centering
\renewcommand{\arraystretch}{1.4}
\caption{\textbf{Temperature scaling is split-robust.} Over $20$ random $50/50$
fit/eval splits: optimal temperature $T$ (mean$\pm$sd), held-out ECE before and after scaling, and the
per-split fold-reduction ECE$_{\text{pre}}$/ECE$_{\text{post}}$. ECE pre/post are on the held-out half,
so they differ slightly from \cref{tab:calibration_rigor}.}
\label{tab:recal_rigor}
\small
\setlength{\tabcolsep}{6pt}
\begin{tabular}{@{}lcccc@{}}
\toprule
\textbf{Model} & \textbf{$T$} & \textbf{ECE pre} & \textbf{ECE post} & \textbf{reduction} \\
\midrule
Qwen2.5-32B & $5.9\pm0.3$ & $0.125\pm0.009$ & $0.038\pm0.010$ & $3.5\times$ \\
Gemma-2-9B & $6.2\pm0.3$ & $0.169\pm0.012$ & $0.037\pm0.011$ & $5.0\times$ \\
Qwen2.5-7B & $8.1\pm0.4$ & $0.194\pm0.011$ & $0.051\pm0.015$ & $4.1\times$ \\
Mistral-7B & $5.7\pm0.4$ & $0.198\pm0.016$ & $0.035\pm0.011$ & $6.0\times$ \\
OLMo-2-7B & $2.7\pm0.2$ & $0.125\pm0.013$ & $0.048\pm0.012$ & $2.8\times$ \\
Llama-3.1-8B & $9.0\pm1.2$ & $0.368\pm0.015$ & $0.118\pm0.020$ & $3.2\times$ \\
\bottomrule
\end{tabular}
\end{table}

%% file: tables/tab_guard.tex
\begin{table}[H]
\centering
\renewcommand{\arraystretch}{1.4}
\caption{\textbf{General-purpose LLMs vs.\ dedicated guards} on identical items, labels, and
metrics (mean over the guard-scored benchmarks; Ethics excluded as out-of-construct; each guard via its
native interface). The general-LLM row is the mean over the general models; \emph{overlap} marks a guard
scored on a benchmark in its own training data.}
\label{tab:guard}
\small
\setlength{\tabcolsep}{6pt}
\begin{tabular}{@{}lcccc@{}}
\toprule
\textbf{Model} & \textbf{FA}$\downarrow$ & \textbf{Miss}$\downarrow$ & \textbf{ECE}$\downarrow$ & \textbf{AURC}$\downarrow$ \\
\midrule
General-purpose LLMs (mean) & 0.329 & 0.182 & 0.205 & 0.140 \\
\midrule
Llama Guard 3 8B & 0.048 & 0.345 & 0.155 & 0.094 \\
WildGuard 7B & 0.131 & 0.109 & 0.092 & 0.069 \\
ShieldGemma 9B & 0.038 & 0.502 & 0.178 & 0.174 \\
Aegis 7B & 0.187 & 0.167 & 0.076 & 0.083 \\
\bottomrule
\end{tabular}
\end{table}

%% file: tables/tab_confsignal.tex
\begin{table}[H]
\centering
\renewcommand{\arraystretch}{1.4}
\caption{\textbf{Confidence signals for error ranking.} AURC (lower is better) when
abstention is ranked by each signal: token-logprob, verbalized $1$--$10$, and sampled-answer agreement.
(Predictive entropy is omitted: for a binary decision it is monotone in logprob confidence and gives an
identical ranking.)}
\label{tab:confsignal}
\small
\setlength{\tabcolsep}{6pt}
\begin{tabular}{@{}lccc@{}}
\toprule
\textbf{Model} & \textbf{Logprob} & \textbf{Verbalized} & \textbf{Sample-agree} \\
\midrule
Qwen2.5-32B & 0.048 & 0.067 & 0.115 \\
Gemma-2-9B & 0.082 & 0.097 & 0.163 \\
Qwen2.5-7B & 0.096 & 0.102 & 0.190 \\
Mistral-7B & 0.137 & 0.152 & 0.232 \\
OLMo-2-7B & 0.246 & 0.305 & 0.304 \\
Llama-3.1-8B & 0.300 & 0.329 & 0.427 \\
\bottomrule
\end{tabular}
\end{table}

%% file: tables/tab_calibration.tex
\begin{table}[H]
\centering
\renewcommand{\arraystretch}{1.4}
\caption{\textbf{Selective risk under abstention.} \emph{Risk@$c$} is the error rate when the
flagger answers only its most-confident fraction $c$ of items and abstains on the rest, pooled over
benchmarks.}
\label{tab:calibration}
\small
\setlength{\tabcolsep}{6pt}
\begin{tabular}{@{}lccc@{}}
\toprule
\textbf{Model} & \textbf{Risk@1.0} & \textbf{Risk@0.8} & \textbf{Risk@0.5} \\
\midrule
Qwen2.5-32B & 0.136 & 0.073 & 0.044 \\
Gemma-2-9B & 0.173 & 0.126 & 0.084 \\
Qwen2.5-7B & 0.205 & 0.137 & 0.084 \\
Mistral-7B & 0.255 & 0.209 & 0.120 \\
OLMo-2-7B & 0.322 & 0.290 & 0.247 \\
Llama-3.1-8B & 0.464 & 0.433 & 0.332 \\
\bottomrule
\end{tabular}
\end{table}

%% file: tables/tab_abstention_controls.tex
\begin{table}[t]
\centering
\renewcommand{\arraystretch}{1.4}
\caption{\textbf{Confidence-based abstention beats random.} Selective risk at
$50\%$ coverage under \emph{random} and \emph{confidence}-based abstention; \emph{Gain}=random$-$confidence.
Every model gains, so the improvement is not merely from answering fewer items.}
\label{tab:abstain_ctrl}
\small
\setlength{\tabcolsep}{6pt}
\begin{tabular}{@{}lccc@{}}
\toprule
\textbf{Model} & \textbf{Random} & \textbf{Confidence} & \textbf{Gain} \\
\midrule
Qwen2.5-32B & 0.136 & 0.044 & 0.092 \\
Gemma-2-9B & 0.173 & 0.084 & 0.089 \\
Qwen2.5-7B & 0.205 & 0.084 & 0.120 \\
Mistral-7B & 0.255 & 0.120 & 0.135 \\
OLMo-2-7B & 0.322 & 0.247 & 0.074 \\
Llama-3.1-8B & 0.464 & 0.332 & 0.132 \\
\bottomrule
\end{tabular}
\end{table}

%% file: tables/tab_abstention_controls_rigor.tex
\begin{table}[H]
\centering
\renewcommand{\arraystretch}{1.4}
\caption{\textbf{Pooled vs.\ within-benchmark abstention.} Selective risk at $50\%$ coverage under
three ranking rules: \emph{pooled} (rank all items globally), \emph{within} (per benchmark, then
averaged), and \emph{norm.} (z-scored within benchmark, then pooled); \emph{gap} is within$-$pooled,
and macro-AURC sits beside pooled AURC.}
\label{tab:abstain_ctrl_rigor}
\small
\setlength{\tabcolsep}{4pt}
\begin{tabular}{@{}lcccccc@{}}
\toprule
 & \multicolumn{4}{c}{\textbf{Selective risk @ 50\% coverage}} & \multicolumn{2}{c}{\textbf{AURC}} \\
\cmidrule(lr){2-5}\cmidrule(lr){6-7}
\textbf{Model} & Pooled & Within & Norm. & Gap & Pooled & Macro \\
\midrule
Qwen2.5-32B & 0.044 & 0.049 & 0.112 & +0.004 & 0.048 & 0.047 \\
Gemma-2-9B & 0.084 & 0.074 & 0.146 & -0.010 & 0.082 & 0.080 \\
Qwen2.5-7B & 0.084 & 0.097 & 0.176 & +0.013 & 0.096 & 0.103 \\
Mistral-7B & 0.120 & 0.122 & 0.116 & +0.001 & 0.137 & 0.139 \\
OLMo-2-7B & 0.247 & 0.193 & 0.209 & -0.054 & 0.246 & 0.195 \\
Llama-3.1-8B & 0.332 & 0.299 & 0.296 & -0.033 & 0.300 & 0.290 \\
\bottomrule
\end{tabular}
\end{table}

%% file: tables/tab_robustness.tex
\begin{table}[H]
\centering
\renewcommand{\arraystretch}{1.4}
\caption{\textbf{Prompt sensitivity of error direction.} Error direction (and false-alarm/miss
rates) for four models on three benchmarks under the base prompt, a \emph{label-order swap} (the
letters for flag / do-not-flag are exchanged), and a \emph{policy paraphrase} (instruction reworded,
order unchanged). Cells where the direction flips relative to the base prompt are in \textbf{bold}.}
\label{tab:robustness}
\scriptsize
\setlength{\tabcolsep}{4pt}
\begin{tabular}{@{}llccc@{}}
\toprule
\textbf{Model} & \textbf{Benchmark} & \textbf{Base} & \textbf{Swap} & \textbf{Paraphrase} \\
\midrule
Llama-3.1-8B & XSTest    & over (0.97 / 0.00) & over (0.33 / 0.03) & over (0.49 / 0.01) \\
Llama-3.1-8B & Ethics    & over (1.00 / 0.00) & over (0.42 / 0.17) & over (0.93 / 0.03) \\
Llama-3.1-8B & WildGuard & over (0.68 / 0.21) & over (0.36 / 0.12) & over (0.45 / 0.24) \\
Qwen2.5-7B   & XSTest    & over (0.21 / 0.00) & over (0.16 / 0.03) & over (0.24 / 0.00) \\
Qwen2.5-7B   & Ethics    & under (0.12 / 0.47) & under (0.20 / 0.25) & under (0.12 / 0.51) \\
Qwen2.5-7B   & WildGuard & under (0.12 / 0.27) & under (0.06 / 0.33) & under (0.13 / 0.30) \\
Gemma-2-9B   & XSTest    & over (0.34 / 0.00) & over (0.14 / 0.00) & over (0.35 / 0.00) \\
Gemma-2-9B   & Ethics    & over (0.26 / 0.06) & \textbf{under (0.04 / 0.46)} & over (0.30 / 0.06) \\
Gemma-2-9B   & WildGuard & over (0.25 / 0.12) & \textbf{under (0.06 / 0.30)} & over (0.30 / 0.07) \\
OLMo-2-7B    & XSTest    & under (0.09 / 0.16) & under (0.05 / 0.25) & \textbf{over (0.14 / 0.04)} \\
OLMo-2-7B    & Ethics    & under (0.03 / 0.80) & \textbf{over (0.73 / 0.02)} & under (0.06 / 0.50) \\
OLMo-2-7B    & WildGuard & under (0.05 / 0.33) & \textbf{over (0.31 / 0.09)} & under (0.17 / 0.24) \\
\bottomrule
\end{tabular}
\end{table}